\documentclass[10pt,twocolumn,letterpaper]{article}

\usepackage[pagenumbers]{iccv} 

\definecolor{iccvblue}{rgb}{0.21,0.49,0.74}
\definecolor{lightblue}{rgb}{0.8, 0.8, 1}
\usepackage[pagebackref,breaklinks,colorlinks,allcolors=iccvblue]{hyperref}
\usepackage{xcolor}
\usepackage{colortbl}

\def\paperID{1177}
\def\confName{ICCV}
\def\confYear{2025}

\title{Next-Scale Autoregressive Models are\\Zero-Shot Single-Image Object View Synthesizers}

\author{Shiran Yuan\\
UC Berkeley\\
{\tt\small shiran\_yuan@berkeley.edu}
\and
Hao Zhao\\
Tsinghua University\\
{\tt\small zhaohao@air.tsinghua.edu.cn}
}

\begin{document}
\maketitle

\begin{abstract}
Methods based on diffusion backbones have recently revolutionized novel view synthesis (NVS). However, those models require pretrained 2D diffusion checkpoints (\emph{e.g.}, Stable Diffusion) as the basis for geometrical priors. Since such checkpoints require exorbitant amounts of data and compute to train, this greatly limits the scalability of diffusion-based NVS models. We present Next-Scale Autoregression Conditioned by View (ArchonView), a method that significantly exceeds state-of-the-art methods despite being trained from scratch with 3D rendering data only and no 2D pretraining. We achieve this by incorporating both global (pose-augmented semantics) and local (multi-scale hierarchical encodings) conditioning into a backbone based on the next-scale autoregression paradigm. Our model also exhibits robust performance even for difficult camera poses where previous methods fail, and is several times faster in inference speed compared to diffusion. We experimentally verify that performance scales with model and dataset size, and conduct extensive demonstration of our method's synthesis quality across several tasks. Our code is open-sourced at \url{https://github.com/Shiran-Yuan/ArchonView}.
\end{abstract}

\begin{figure*}
\centering
\includegraphics[width=\textwidth]{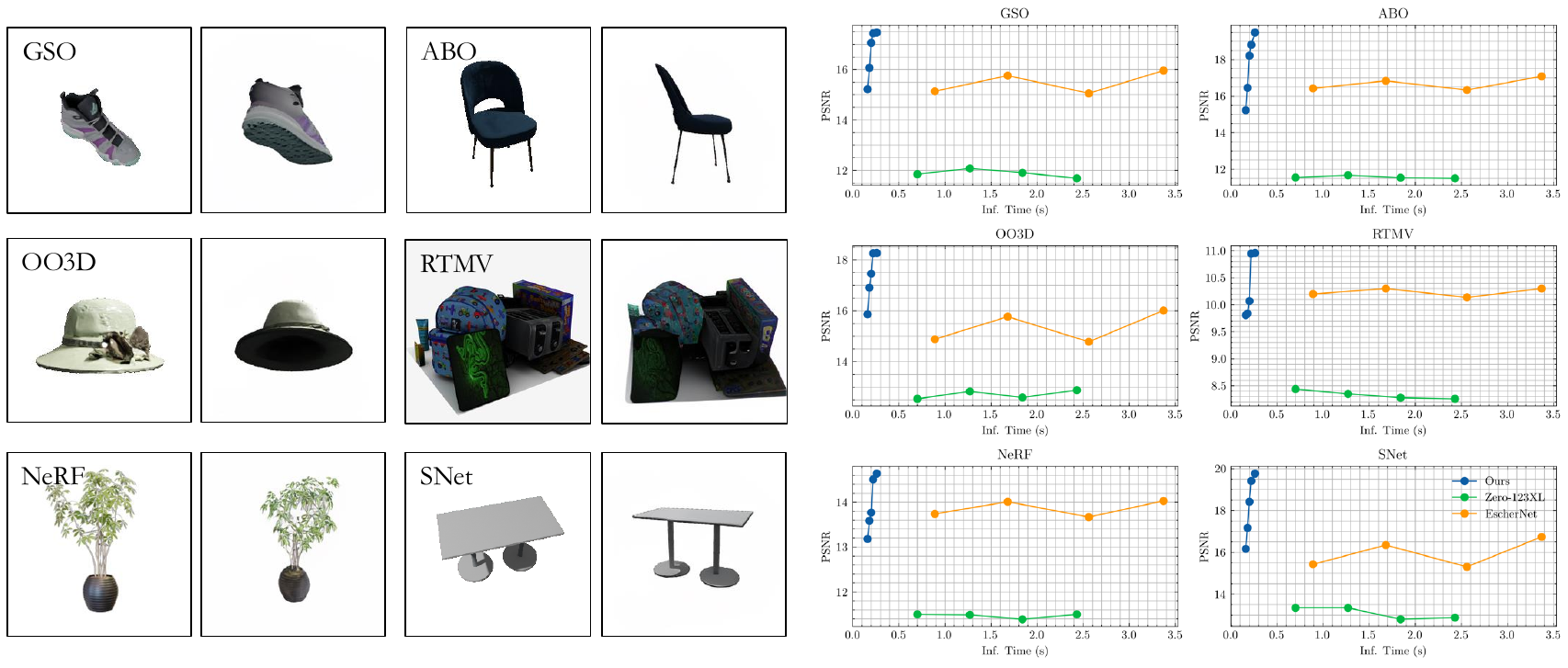}
\caption{\textbf{Fast, accurate, and scalable novel view synthesis without 2D pretraining.} Displayed on the left are results sampled from each of our test datasets, where the left image shows the input and the right image is a novel view from our model. Note that all of the displayed results are from zero-shot inference. On the right are time \emph{v.s.} PSNR tradeoff plots, where our models scaled to different parameter sizes are compared against diffusion models with different denoising steps.}
\label{fig:teaser}
\vspace{-2ex}
\end{figure*}

\section{Introduction}
Humans, living in a 3D world, naturally infer the complete 3D structure of objects from a single 2D view, leveraging prior knowledge and spatial reasoning. If machines could achieve the same, particularly in a zero-shot manner for unseen objects, it would greatly benefit fields such as 3D content creation, simulation, and real-world perception systems. Consequently, zero-shot novel view synthesis (NVS) from single object-centric images emerges as a fundamental challenge in computer vision. Since this is a highly under-constrained problem, it is typically formulated as a generative task conditioned on the input image and relative camera pose. The prevailing approach fine-tunes a 2D diffusion model to exploit implicit geometric priors learned from large-scale image datasets. Whilst this paradigm has shown promising results, it comes with several limitations.

The primary limitation of diffusion-based NVS models lies in their reliance on pretrained 2D diffusion backbones, which require vast amounts of data and computational resources—far beyond what is accessible to most researchers. This fundamental constraint makes scaling diffusion-based NVS models extremely challenging. Notably, to the best of our knowledge, no prior work has demonstrated scaling trends with respect to model size for single-image object-centric NVS. Another critical limitation of diffusion models is their inherent trade-off between speed and quality. Due to the need for multiple denoising steps through a U-Net structure, achieving high-quality outputs inevitably results in relatively slow inference. Consequently, even if diffusion models were scalable, larger models would further increase inference time and computational cost, potentially rendering them impractical for real-world deployment.

Such downsides of diffusion call for a paradigm shift: a backbone which can be readily scaled up, does not require 2D pretraining, is more efficient, and outputs better results. To this end we propose Autoregression Conditioned by View (ArchonView), the first NVS model based on visual autoregressive generation. We base our method on the recently proposed next-scale autoregression backbone, which replaces raster-scan-ordered next-token prediction in conventional visual autoregression with autoregressively predicting the next resolution scale of tokens in a coarse-to-fine manner (described in Sec.~\ref{sec:ar}). Whilst previous works have verified its scalability and applicability to many tasks in vision, so far, its applicability to NVS remains unknown.

In order to adapt autoregression to NVS, we condition the model in two ways: globally, through a posed CLIP encoding, which allows the model to gain semantic information augmented by the desired relative pose (described in Sec.~\ref{sec:global}); and locally, encoded by the multiscale VQVAE representation (which is also used for quantization in the main generative architecture), enforcing consistency in local details with the input image (described in Sec.~\ref{sec:local}). Experimentation shows that our method achieves state-of-the-art performance consistently and robustly across multiple benchmarks, scales with both model size and dataset size, and is several times faster than current diffusion-based methods. Some of our results across different evaluation datasets and comparisons with current methods are shown in Fig.~\ref{fig:teaser}.

In summary, our key technical contributions to the field of zero-shot single-image object NVS are as follows:
\begin{itemize}
\item We present a method that does not require any pretraining on 2D data, thereby drastically lowering the computation and data requirement for training a model from scratch.
\item Our method consistently achieves state-of-the-art performance and is also several times faster in terms of inference time compared to previous works.
\item We demonstrate that our method scales with both model size and dataset size.
\item We are the first to base a model on an autoregressive backbone for this task, demonstrating the potentials of next-scale autoregression.
\end{itemize}

\section{Related Works}
\begin{figure*}
\centering
\includegraphics[width=\textwidth]{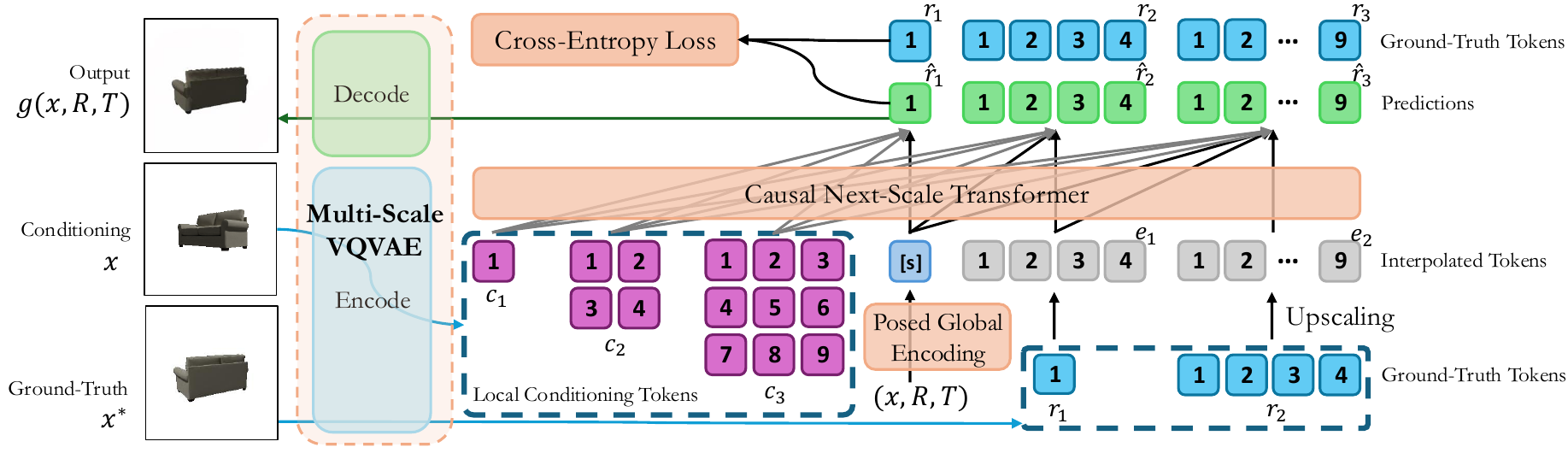}
\vspace{-4ex}
\caption{\textbf{The overall architecture for the training of ArchonView.} The predictions are classifier logit predictions, and can be converted to images after sampling based on the logit probabilities. The loss calculation is directly based on the logits and does not involve sampling.}
\label{fig:training}
\vspace{-2ex}
\end{figure*}

\subsection{Generative Modeling}
Autoregression has seen many applications to generation of language~\cite{brown2020language, radford2018improving, radford2019language, touvron2023llama}, world models~\cite{bruce2024genie, lu2024genex, tu2025videoanydoor}, videos~\cite{kondratyuk2024videopoet, wu2022nuwa, yan2021videogpt}, and multimodal outputs~\cite{chameleon, kelly2024visiongpt, gpt4v, sun2023emu}. Besides being efficient and accurate enough for operational use, it has also been shown to be scalable in many tasks~\cite{henighan2020scaling, kaplan2020scaling}. However, the current predominant paradigm of 2D generative modeling is indubitably diffusion, thanks to multiple groundbreaking innovations in this field~\cite{ho2021classifier, peebles2023scalable, rombach2022high, zhang2023adding}. Meanwhile, though conventional visual autoregression (using a raster-scan traversal of fixed-size patches as tokens) has produced innovative techniques~\cite{esser2021taming, lee2022autoregressive, parmar2018image, van2017neural, yu2022vector} and achieved some milestones~\cite{chen2020generative, ramesh2021zero, razavi2019generating, yu2022scaling}, advances in diffusion left it largely irrelevant.

Recently, the newly proposed next-scale prediction paradigm of autoregression~\cite{tian2024visual}, replacing the conventional raster-scan next-token paradigm, has been proven effective, developed upon~\cite{gu2024dart, ren2024flowar, ren2024m, tang2025hart}, and adapted~\cite{han2024infinity, li2024controlar, ma2024star, yao2024car, zhang2024var}. Empirical evidence demonstrates its reliability in achieving accurate, efficient, and scalable results, exceeding diffusion models in many tasks where raster-scan autoregression struggles. This has reignited interest in using autoregressive models for visual generative tasks where diffusion models currently dominate.

\subsection{Novel View Synthesis}
Before the advent of generative NVS models, predominant methodologies have used implicit representations~\cite{barron2021mip, barron2022mip, barron2023zip, mildenhall2021nerf}, voxel-like representations~\cite{chen2022tensorf, fridovich2022plenoxels, sun2022direct, yu2021plenoctrees}, explicit primitives~\cite{huang20242d, kerbl20233d, mai2024ever, muller2022instant}, \emph{etc.} to model 3D scenes or objects based on given views and poses, thus achieving NVS. However, in the case of sparse inputs, where few views or only one view is available, none of those models are able to produce accurate results due to the scene being severely underconstrained. 

While some efforts have been made to adapt conventional frameworks to sparse-input scenarios~\cite{chen2021mvsnerf, chibane2021stereo, jain2021putting, niemeyer2022regnerf, xu2022sinnerf, yu2021pixelnerf}, their capabilities were limited, or often depended on strict hypotheses regarding geometrical priors. In addition, most aforementioned methods require specific training or fine-tuning on the scene or object in question, or were fine-tuned on a class of objects (\emph{e.g.}, from ShapeNet~\cite{chang2015shapenet}) and only perform well for in-distribution inputs. Thus, none achieved capability for zero-shot single-image NVS with such paradigms.

\subsection{Generative NVS}
Early works have used GANs~\cite{goodfellow2014generative} as backbones for conducting NVS generatively, reframing the problem as modeling the distribution of scene views conditioned by camera pose~\cite{chan2021pi, chan2022efficient, gadelha20173d, nguyen2019hologan, niemeyer2021giraffe, schwarz2022voxgraf}. In late 2022 through 2023 there was a surge of works using diffusion models for NVS, often coupled with the then-recent NVS representations (\emph{e.g.} NeRF); in particular, some works used diffusion models as priors for supervising the training of a 3D representation model~\cite{bautista2022gaudi, deng2023nerdi, melas2023realfusion, sargent2024zeronvs, wang2023sparsenerf, wu2024reconfusion}, while others directly used diffusion models with camera pose conditioning as an NVS representation backbone~\cite{chan2023generative, liu2023zero, watson2023novel} (interestingly the latter's methodology of view generation based on latents coincides with previous works using transformers~\cite{kulhanek2022viewformer, rombach2021geometry, sajjadi2022scene}).

A particularly important work in this period was Zero 1-to-3~\cite{liu2023zero}, which was fine-tuned on an image variation model~\cite{pinkney2023stable} which, in turn, was tuned on Stable Diffusion~\cite{rombach2022high}. Its main conclusion was that 2D diffusion models already contain 3D-aware priors, and that such priors can be directly extracted by fine-tuning pretrained 2D diffusion checkpoints. This paradigm of fine-tuning 2D diffusion models for priors spawned a line of works~\cite{kong2024eschernet, liu2024syncdreamer, shi2023zero123++, weng2024consistent123, ye2024consistent, zheng2024free3d} which similarly attempt to extract 3D-aware priors from pretrained 2D diffusion models. In our work, we demonstrate that autoregressive models can directly have such 3D-awareness without 2D pretraining.

\section{Methodology}

\subsection{Motivation}
\paragraph{Problem Formulation}
Formally, we wish to solve the following problem. Given a single input view $x$ of an underlying 3D object, as well as relative camera transformations $R\in\mathbb R^{3\times3}$ and $T\in\mathbb R^3$, we would like to create a probabilistic model $g(x,R,T)$ such that the output view
\begin{equation}
x^*_{R,T}\sim g(x,R,T)
\end{equation}
follows the distribution of the transformed view from applying the relative camera transformation $(R,T)$ on $x$. 

\paragraph{Diffusion is Not All You Need}
The current predominant formulation of our problem is as a diffusion model. Specifically, common architectures based on the latent diffusion paradigm~\cite{rombach2022high} are made up of an image encoder-decoder pair $(\mathcal E,\mathcal D)$, a U-Net denoiser $\epsilon_\theta$, and a conditioning encoder $\tau$. The latents $z\sim\mathcal E(x)$ are then corrupted with additive Gaussian noise at each step, forming noised latents $z_t$. Diffusion models are thus trained based on the objective
\begin{equation}
\min\limits_\theta\mathbb E_{z\sim\mathcal E(x),\epsilon\sim\mathcal N(0,1),t}||\epsilon-\epsilon_\theta(z_t,t,\tau(x,R,T))||_2^2.
\end{equation}

However, we notice a key downside of this formulation; zero-shot capabilities in this line of research are dependent upon priors within 2D pretrained diffusion models. For instance, one can compare~\cite{watson2023novel} with~\cite{liu2023zero}, which are both based on the presented model; the former was not tuned from 2D models and thus did not exhibit zero-shot abilities, while the latter was tuned from Stable Diffusion and used CLIP for conditioning encoding, thus allowing zero-shot NVS. This naturally leads to a scalability issue: since the underlying checkpoints require large amounts of data and computation to train, it is difficult to scale existing models up. To the best of our knowledge, most or potentially all works in this direction are tuned from Stable Diffusion. Diffusion models also have other problems such as relatively slow inference speed (due to repeated denoising passes), which further limit their effectiveness.

\subsection{Backbone Architecture}
\label{sec:ar}
\begin{figure}
\centering
\includegraphics[width=.45\textwidth]{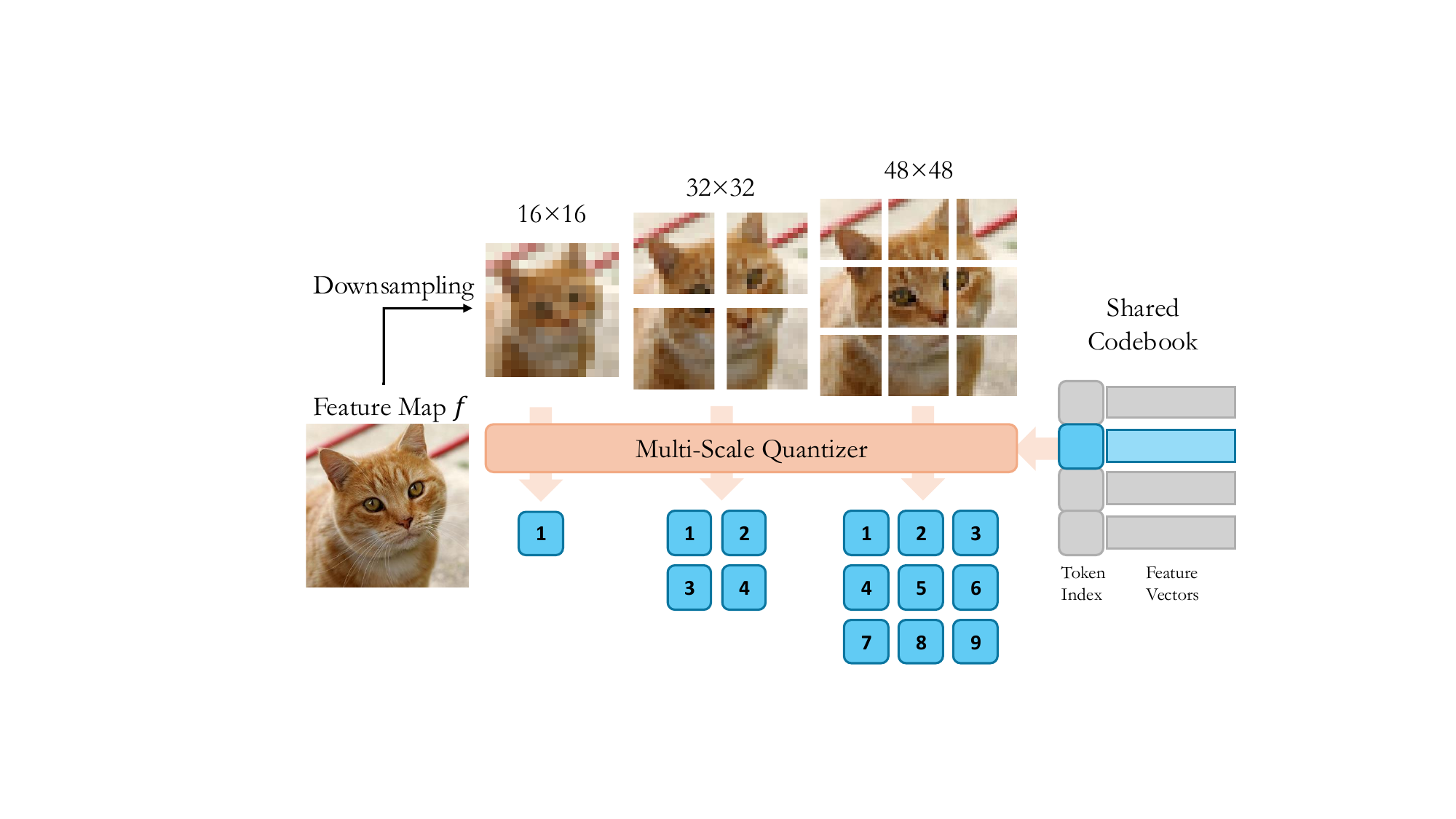}
\caption{\textbf{Structure of the multi-scale VQVAE.} The VQVAE operates by converting the input image into a feature map, resizing the feature map into different scales, and using a codebook shared between scales to compress the patch tokens. Note here $f$ is aesthetically represented as an image whereas it is actually an encoded version of the image.}
\label{fig:vqvae}
\vspace{-2ex}
\end{figure}

\begin{figure}
\centering
\includegraphics[width=.45\textwidth]{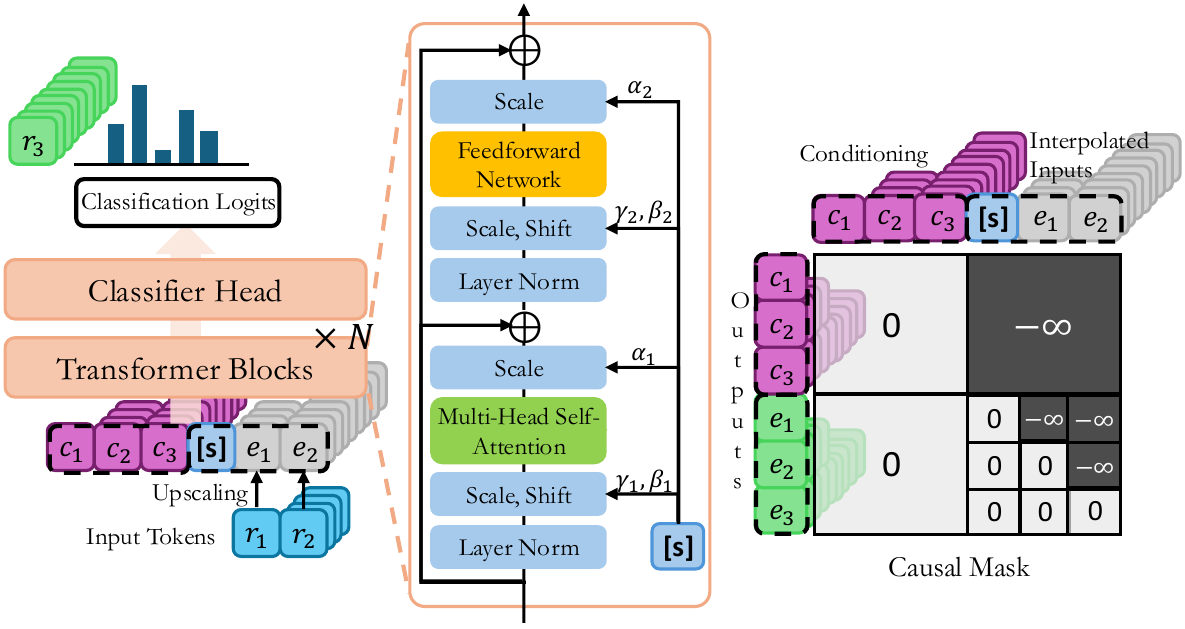}
\caption{\textbf{The next-scale prediction transformer architecture.} Next-scale prediction operates by passing local conditioning tokens and input tokens through multiple transformer blocks which are conditioned by adaptive layer normalization. The transformers use a attention bias mask to enforce the autoregressive causality constraint.}
\label{fig:transformers}
\vspace{-2ex}
\end{figure}

We propose using next-scale autoregression~\cite{tian2024visual} as our backbone paradigm instead and modifying it to suit our task. Our overall architecture during training is shown in Fig.~\ref{fig:training}. Next-scale autoregression is based on predicting the next resolution scale of the image in a coarse-to-fine manner, and consists of two main parts: a multi-scale vector-quantized variational autoencoder (VQVAE)~\cite{van2017neural} which supports image tokenization (depicted in Fig.~\ref{fig:vqvae}), and a transformer which autoregressively predicts the next scale of tokens (depicted in Fig.~\ref{fig:transformers}). 

\paragraph{Multi-Scale Tokenization}
The VQVAE contains an image encoder-decoder pair $(\mathcal E,\mathcal D)$, a quantizer $\mathcal Q$, and a learnable codebook $Z$ with vocabulary size $V$. The image encoder takes an image $x$ as input and encodes it into a feature map $f=\mathcal E(x)\in\mathbb R^{h\times w\times C}$ (where $h$,$w$ are the latent height/width, and $C$ is the embedding dimension). Note that the latent dimensions of the image are based on patching, similar to conventional autoregression and latent diffusion. The feature maps are then quantized as $q=\mathcal Q(f)\in[1..V]^{h\times w}$. The quantizer does this by mapping each patch $f_{i,j}$ to the Euclidean nearest codebook entry $Z_v$:
\begin{equation}
q_{i,j}=\mathop{\arg\min}\limits_{v\in[1..V]}||Z_v-f_{i,j}||_2.
\end{equation}
The key to the multi-scale formulation is that the input image is resized into different resolution scales. For instance, since we use $16\times16$ patches, the $3\times3$ scale would reshape the image to $48\times48$ pixel resolution before tokenization; and the corresponding scale token would contain 9 patch tokens. All scales share the same codebook $Z$, thus ensuring a consistent vocabulary is used across different token scales. The image can then be reconstructed using the decoder $\mathcal D$ given the quantized tokens. 

\paragraph{Next-Scale Prediction Transformers}
Based on a frozen VQVAE (trained on a compound loss as in~\cite{tian2024visual}), the autoregressive model is formulated as follows. We attempt to achieve a probabilistic model $p_
\theta$ (the next-scale transformer) conditioned by our inputs $(x,R,T)$ such that the joint likelihood of scale tokens $(r_1,r_2,\cdots,r_K)$ representing the distribution of $x_{R,T}^*$ can be modeled as
\begin{align}
\label{eq:ar}
g(x,R,T)&=p_\theta(r_1,r_2,\cdots,r_K|x,R,T)\\\nonumber
&=\prod\limits_{k=1}^Kp_\theta(r_k|x,R,T,r_1,r_2,\cdots,r_{k-1}).
\end{align}
Note that here the autoregressive assumption is that each scale $r_k$ only depends on the conditioning and the previous scales (similar to a coarse-to-fine process), instead of the later scales. We add conditioning tokens $c_k$ (details in Sec.~\ref{sec:local}) as well for additional information, and the causal mask used for transformers is as displayed in Fig.~\ref{fig:transformers}. 

We use teacher forcing to train the model, where $(r_1,r_2,\cdots,r_{K-1})$ of $x^*$ and conditioning from $x$ are used for causally predicting all tokens $(r_1,r_2,\cdots,r_K)$ of $x^*$. After sending inputs and conditioning through several transformer blocks, the eventual results are passed through a classifier head (consisting of a single linear layer) and outputted as classification logits, after which the cross-entropy loss between the logits and ground truths is taken for backpropagation. 

Implementation-wise, each scale token $r_k$ is first interpolated to the same size as $r_{k+1}$ by taking the corresponding feature map $f$, resizing it accordingly, and applying tokenization. An exception is the $[s]\to r_1$ mapping, which does not require resizing. The upscaled results $e_k$ are then taken as input to the causal transformer. Each attention layer uses adaptive layer normalization (AdaLN)~\cite{peebles2023scalable} conditioned by the start token (to ensure consistency with the global encoding conditioning) and multi-head self-attention~\cite{vaswani2017attention}. We refer to the number of AdaLN transformer blocks per prediction step as the model depth and the dimension of tokens $C$ as the model width. For easy scaling, in our experiments we set the model width to always be 64 times the depth.

During inference, we first find $c_k$ and the start token [s] using $x$. We then use them to infer the distribution of $r_1$ and sample from the distribution. We then use the available information to infer $r_2$, and so on. Note that already known tokens are kv-cached~\cite{pope2023efficiently} and not replaced in further inference steps. The final inferred $r_k$ values are passed through the VQVAE decoder to arrive at $\hat x^*$.

\paragraph{The Devil is in the Classifier Head}
The original formulation of the next-scale transformer~\cite{tian2024visual} applies adaptive normalization before the classifier head as well. However, as we will demonstrate in experiments, in our task this restricts the performance of the resulting model considerably. We theorize that this implies global conditioning using AdaLN must be accompanied by self-attention with local conditioning tokens (as the classification logit prediction step does not involve attention transformers), and that otherwise only using global conditioning would prompt the output image to ``forget'' details from local conditioning. After the removal of this normalization step, our results improved considerably.

\subsection{Semantic Global Pose Conditioning}
\label{sec:global}
We now start presenting the details of how we adapted the architecture via adding conditioning for our task. Firstly, we need to choose a start token that reliably captures global information from our conditioning $(x,R,T)$, because it will be used both for initializing the autoregressive procedure and for normalization of every attention layer, which means it should have complete field of view. Meanwhile, it must also be a single token because it will be mapped to the first resolution scale (which comprises of a single patch) during autoregressive inference. 

To balance those two needs, we would like an encoding scheme that could condense semantic information from the image along with the query poses into a vector of size comparable to tokens. Inspired by~\cite{liu2023zero}, we use a ``Posed CLIP'' embedding as follows:
\begin{equation}
\tau(x,R,T)=W(W_i(\textrm{CLIP}(x)\oplus[\theta,\sin\phi,\cos\phi,r])+b_i)+b.
\end{equation}
Here, $\theta$, $\phi$, and $r$ respectively stand for the relative elevation, azimuth, and radial distance of the transformation; $\oplus$ stands for concatenation along the feature dimension; $\textrm{CLIP}(x)\in\mathbb R^{768}$ is the CLIP visual embedding~\cite{radford2021learning}, which we use here as a global semantic encoder; and $(W_i,b_i),(W,b)$ are two pairs of linear layer parameters, where the $i$ subscript stands for identity initialization (\emph{i.e.} $W_i$ is initialized as a generalized identity matrix and $b$ is initialized as a zero vector).

We add classifier-free guidance by randomly replacing values in the 768-dimensional posed CLIP embedding (which has gone through $(W_i,b_i)$ but not $(W,b)$ yet) with values from a null CLIP text embedding (applying the CLIP text encoder to an empty string). 

The two linear layers here serve different purposes. The identity-initialized layer $W_i\in\mathbb R^{768\times772},b_i\in\mathbb R^{768}$ aims to map the relative pose information into the original 768 CLIP features (which is why initially the layer outputs the CLIP embedding with no regards to the camera pose). The other layer $W\in\mathbb R^{C\times768},b\in\mathbb R^{C}$, initialized normally, serves as an interface between the semantic embedding and the transformer architecture by mapping it onto a token. The $(W_i,b_i)$ layer is set to have 10 times the learning rate of other parameters, as in our experiments without this setting the gradient quickly explodes.

\begin{table*}
\centering
\footnotesize
\caption{\textbf{Quantitative benchmarking results.} We compare against state-of-the-art baselines using diffusion backbones on six well-established benchmarking datasets.}
\begin{tabular}{ccccccccccc}
\toprule
 & & \textbf{GSO} & & & \textbf{ABO} & & & \textbf{OO3D} \\
\cmidrule(lr){2-4}\cmidrule(lr){5-7}\cmidrule(lr){8-10}
 & PSNR ($\uparrow$) & SSIM ($\uparrow$) & LPIPS ($\downarrow$) & PSNR ($\uparrow$) & SSIM ($\uparrow$) & LPIPS ($\downarrow$) & PSNR ($\uparrow$) & SSIM ($\uparrow$) & LPIPS ($\downarrow$) & Time (s)\\
Zero 1-to-3~\cite{liu2023zero} & 13.39 & 0.7776 & 0.2672 & 12.75 & 0.7632 & 0.2901 & 13.43 & 0.7737 & 0.2723 & 1.84 \\
Zero 123-XL~\cite{deitke2023objaverse-xl} & 13.80 & 0.7865 & 0.2595 & 12.78 & 0.7646 & 0.2766 & 14.05 & 0.7966 & 0.2516 & 1.84\\
EscherNet~\cite{kong2024eschernet} & 16.77 & 0.8275 & 0.1891 & 17.03 & 0.8381 & 0.1693 & 16.12 & 0.8294 & 0.2060 & 1.68\\
\cellcolor{lightblue}Ours & \cellcolor{lightblue}\textbf{17.44} & \cellcolor{lightblue}\textbf{0.8491} & \cellcolor{lightblue}\textbf{0.1853} & \cellcolor{lightblue}\textbf{18.82} & \cellcolor{lightblue}\textbf{0.8725} & \cellcolor{lightblue}\textbf{0.1360} & \cellcolor{lightblue}\textbf{18.26} & \cellcolor{lightblue}\textbf{0.8622} & \cellcolor{lightblue}\textbf{0.1617} & \cellcolor{lightblue}\textbf{0.22}\\
\midrule
 & & \textbf{RTMV} & & & \textbf{NeRF} & & & \textbf{SNet} \\
\cmidrule(lr){2-4}\cmidrule(lr){5-7}\cmidrule(lr){8-10}
 & PSNR ($\uparrow$) & SSIM ($\uparrow$) & LPIPS ($\downarrow$) & PSNR ($\uparrow$) & SSIM ($\uparrow$) & LPIPS ($\downarrow$) & PSNR ($\uparrow$) & SSIM ($\uparrow$) & LPIPS ($\downarrow$) & Params\\
Zero 1-to-3~\cite{liu2023zero} & 8.49 & 0.5260 & 0.4772 & 10.88 & 0.6222 & 0.4146 & 13.02 & 0.7957 & 0.3288 & 1.0B \\
Zero 123-XL~\cite{deitke2023objaverse-xl} & 8.58 & 0.5237 & 0.4735 & 11.29 & 0.6551 & 0.3926 & 13.29 & 0.8070 & 0.3206 & 1.0B\\
EscherNet~\cite{kong2024eschernet} & 10.38 & 0.5327 & 0.4340 & 13.85 & 0.6783 & 0.2868 & 16.35 & 0.8450 & 0.1951 & 1.0B\\
\cellcolor{lightblue}Ours & \cellcolor{lightblue}\textbf{10.95} & \cellcolor{lightblue}\textbf{0.5739} & \cellcolor{lightblue}\textbf{0.3991} & \cellcolor{lightblue}\textbf{14.51} & \cellcolor{lightblue}\textbf{0.7025} & \cellcolor{lightblue}\textbf{0.2735} & \cellcolor{lightblue}\textbf{19.42} & \cellcolor{lightblue}\textbf{0.8927} & \cellcolor{lightblue}\textbf{0.1300} & \cellcolor{lightblue}\textbf{1.0B}\\
\bottomrule
\end{tabular}
\label{tab:bench}
\end{table*}

\subsection{Multi-Scale Local Conditioning}
\label{sec:local}
The global encoding is a good condition for generation because it aggregates semantic information across the entire image and has full field-of-view. However, in this process, the details of the original image are lost. Hence we need to add another source of conditioning which can directly provide the autoregressive model with portions of the input image. To this end we propose a local conditioning mechanism which exploits our multi-scale tokenizer to provide effective conditioning information for the attention layers.

As previously shown in Fig.~\ref{fig:training}, we encode the conditioning image $x$ through the multi-scale VQVAE into tokens the same way as we encode the ground-truth $x^*$. This ensures that the conditioning tokens and tokens used during generation share the same vocabulary (from the VQVAE codebook $Z$), thus facilitating the application of self-attention in the next-scale transformer. We extend the block triangular mask such that all conditioning tokens can affect the next attention block's conditioning and the input tokens, but are not affected by input tokens (as displayed in Fig.~\ref{fig:transformers}). This method of conditioning via token prepending is well-suited to the autoregressive nature of our paradigm, and does not require any architectural accommodations. We observe in our ablation studies that this greatly enhances the effectiveness of our method compared to purely using a global encoding as conditioning. We add classifier-free guidance by randomly replacing local conditioning tokens with a learnable null token.

In conventional diffusion-based methods, the predominant way of adding local conditioning is to channel-concatenate the conditioning image onto the latent noise used for diffusion. However, as previous works have noted~\cite{shi2023zero123++}, this inherently creates a false pixel-to-pixel correspondence between the conditioning image and the latent whereas the actual relationship is a lot more non-trivial (as transformations in 3D are involved). 

Our method, in contrast, does not inherently impose any correspondences. Since the multi-scale encoding is inherently hierarchical, not only can the transformer learn correspondence relationships without any structural assumptions, it can also model more complex relationships that require information from several levels of detail. 

\begin{table*}
\centering
\footnotesize
\caption{\textbf{Ablation study results.} We evaluate the necessity of two design choices: adding local conditioning (as in Sec.~\ref{sec:local}), and removing adaptive layer normalization (AdaLN) for the classifier head.}
\begin{tabular}{cccccccccc}
\toprule
 & & \textbf{GSO} & & & \textbf{ABO} & & & \textbf{OO3D} \\
\cmidrule(lr){2-4}\cmidrule(lr){5-7}\cmidrule(lr){8-10}
 & PSNR ($\uparrow$) & SSIM ($\uparrow$) & LPIPS ($\downarrow$) & PSNR ($\uparrow$) & SSIM ($\uparrow$) & LPIPS ($\downarrow$) & PSNR ($\uparrow$) & SSIM ($\uparrow$) & LPIPS ($\downarrow$)\\
Global Only & 13.25 & 0.7959 & 0.2672 & 13.04 & 0.7892 & 0.3116 & 15.03 & 0.8161 & 0.2523 \\
w/ Cls. Head AdaLN & 12.76 & 0.7885 & 0.2510 & 12.80 & 0.7549 & 0.3131 & 12.59 & 0.7832 & 0.2540 \\
\cellcolor{lightblue}Ours & \cellcolor{lightblue}\textbf{17.44} & \cellcolor{lightblue}\textbf{0.8491} & \cellcolor{lightblue}\textbf{0.1853} & \cellcolor{lightblue}\textbf{18.82} & \cellcolor{lightblue}\textbf{0.8725} & \cellcolor{lightblue}\textbf{0.1360} & \cellcolor{lightblue}\textbf{18.26} & \cellcolor{lightblue}\textbf{0.8622} & \cellcolor{lightblue}\textbf{0.1617}\\
\midrule
 & & \textbf{RTMV} & & & \textbf{NeRF} & & & \textbf{SNet} \\
\cmidrule(lr){2-4}\cmidrule(lr){5-7}\cmidrule(lr){8-10}
 & PSNR ($\uparrow$) & SSIM ($\uparrow$) & LPIPS ($\downarrow$) & PSNR ($\uparrow$) & SSIM ($\uparrow$) & LPIPS ($\downarrow$) & PSNR ($\uparrow$) & SSIM ($\uparrow$) & LPIPS ($\downarrow$)\\
Global Only & 8.74 & 0.4968 & 0.5124 & 11.80 & 0.6280 & 0.4176 & 14.52 & 0.8232 & 0.2906\\
w/ Cls. Head AdaLN & 7.39 & 0.4771 & 0.5301 & 10.41 & 0.5958 & 0.4237 & 12.73 & 0.7929 & 0.2739 \\
\cellcolor{lightblue}Ours & \cellcolor{lightblue}\textbf{10.95} & \cellcolor{lightblue}\textbf{0.5739} & \cellcolor{lightblue}\textbf{0.3991} & \cellcolor{lightblue}\textbf{14.51} & \cellcolor{lightblue}\textbf{0.7025} & \cellcolor{lightblue}\textbf{0.2735} & \cellcolor{lightblue}\textbf{19.42} & \cellcolor{lightblue}\textbf{0.8927} & \cellcolor{lightblue}\textbf{0.1300}\\
\bottomrule
\end{tabular}
\label{tab:ablation}
\end{table*}

\section{Experiments}

\begin{figure}
\centering
\includegraphics[width=.45\textwidth]{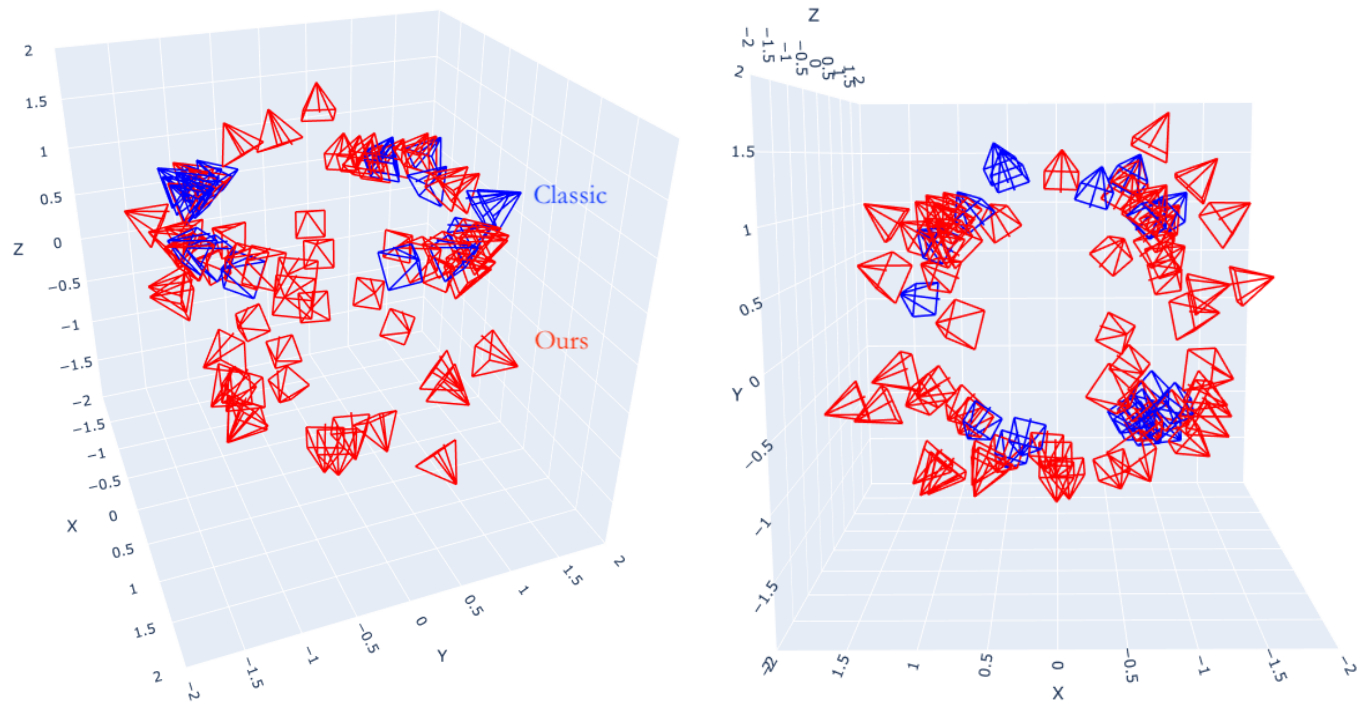}
\caption{\textbf{Improving NVS evaluation.} A free view and a top view of the ``classic'' input camera views compared to ours (40 each sampled).}
\label{fig:cameras}
\vspace{-2ex}
\end{figure}

\begin{figure*}
\centering
\includegraphics[width=\textwidth]{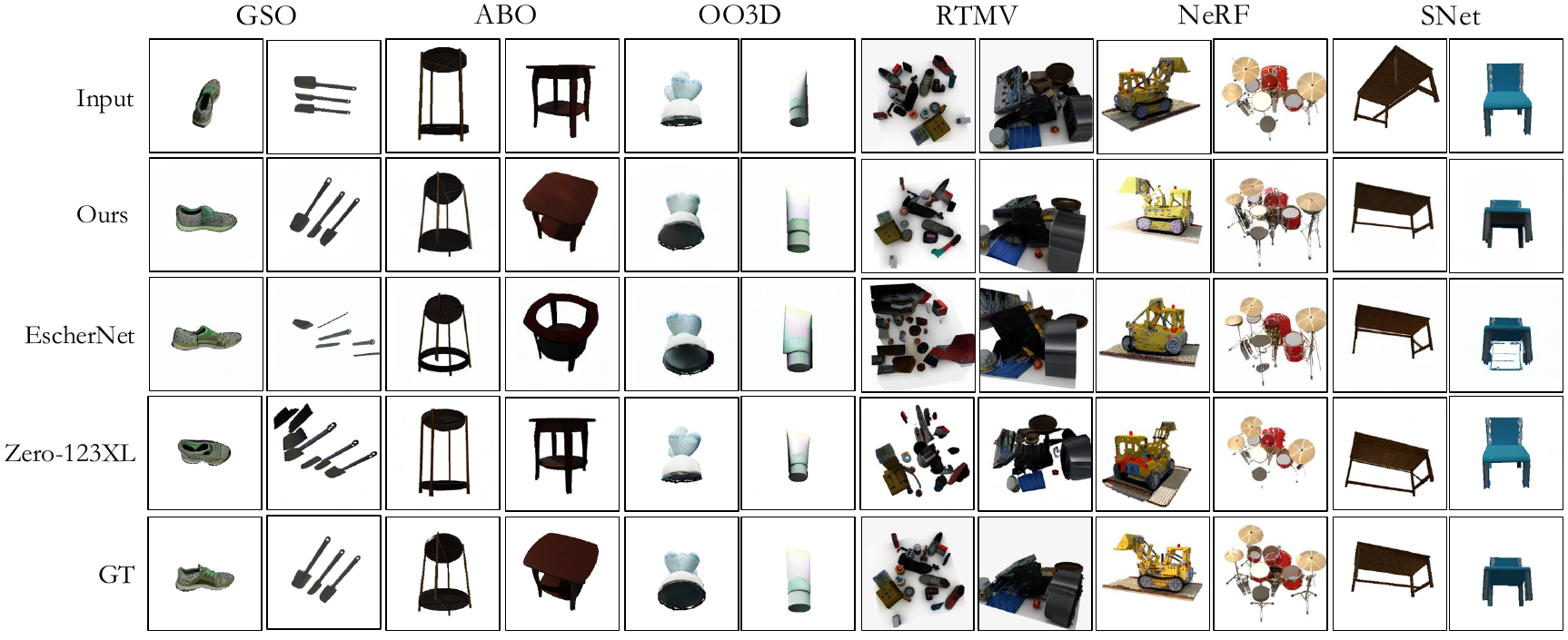}
\caption{\textbf{Qualitative results.} We select a total of 12 cases to visually compare our method with diffusion-based prior works.}
\label{fig:qualitative}
\vspace{-2ex}
\end{figure*}

\subsection{Training Settings}
We use the multi-scale VQVAE checkpoint from~\cite{tian2024visual}, which was trained on the OpenImages dataset~\cite{kuznetsova2020open}. We train our models on the Objaverse dataset~\cite{deitke2023objaverse}, from which we render 256$\times$256 views with randomly sampled camera poses. The number of 16$\times$16 patches per side for each VQVAE scale follows the progression $(1,2,3,4,5,6,8,10,13,16)$, for a total of 10 prediction steps. The depth of the model is set to be 24 for benchmarking in order for the model size to be 1.0B, matching diffusion-based baselines. Optimization is conducted via AdamW~\cite{loshchilov2019decoupled}, and implementation uses PyTorch~\cite{paszke2019pytorch}. We trained on 32 NVIDIA H200 GPUs, and tested inference speed on a single H200. We report baseline results from officially recommended default settings. 

\subsection{Evaluation Settings}
We conduct evaluation across six commonly used datasets: Google Scanned Objects (GSO)~\cite{downs2022google}, Amazon Berkeley Objects (ABO)~\cite{collins2022abo}, OmniObject3D (OO3D)~\cite{wu2023omniobject3d}, Ray-Traced Multi-View Synthetic (RTMV)~\cite{tremblay2022rtmv}, NeRF Synthetic (NeRF)~\cite{mildenhall2021nerf}, and ShapeNet Core (SNet)~\cite{chang2015shapenet}. None of those datasets overlap with our training set (Objaverse), which means that all of the experimental results are zero-shot inference.

For all datasets except NeRF, we sample 100 objects randomly, and render 7 views for 3D objects (for RTMV we directly sample 7 views), with one being used as the input view and the other 6 being used for evaluation; for NeRF, we use all 8 objects, sample one random input view from each object's training set, and evaluate across all 200 testing views.

\paragraph{Improving NVS Benchmarking}
We notice that benchmarking in previous works have usually inherited the evaluation rendering settings used by Zero 1-to-3~\cite{liu2023zero}. However, in this ``classic'' setting, we notice that input views are selected from a narrow range of elevation angles, which makes the task much easier but does not demonstrate the model's robustness across a wide range of possible views of an object. This is also a discrepancy with the training setting, where camera poses are sampled across an entire viewing sphere (save for very high elevation angles, and with the radial distance varying). 

Hence, we choose to make the evaluation setting exactly the same as the training setting. A visual comparison between cameras sampled in our evaluation setting and the ``classic'' one is in Fig.~\ref{fig:cameras}. We note that this makes the task considerably more difficult (but the camera poses are still all within the training range, and hence are not out-of-distribution), and requires models to be more robust with respect to the input pose.

\subsection{Results}

\paragraph{Benchmarking}
Quantitative results from our benchmarking are shown in Tab.~\ref{tab:bench}. As shown, our model, while being of the same size as the Stable Diffusion-tuned baselines, is up to more than 8 times faster than those baselines (under their officially recommended settings), and consistently produces significantly better results across our 6 benchmarks.

\paragraph{Qualitative Comparison}
We present qualitative results in Fig.~\ref{fig:qualitative}. As shown, our model produces more accurate reconstructions, and also can achieve more realistic and feasible results (\emph{e.g.}, GSO shoe) even though it has seen far fewer visual information than the diffusion-based baselines (which have gone through 2D pretraining). Furthermore, our model is also good at localizing objects after transformation even in cluttered scenes (\emph{e.g.}, RTMV), and has a good understanding of geometrical structure (\emph{e.g.}, SNet table). Even when outputs are inaccurate due to uncertainty regarding factors such as lighting (\emph{e.g.}, NeRF lego bulldozer), it continues to present accurate geometries and feasible lighting/texture. Surprisingly, although one might expect diffusion-based models would be more capable of predicting unseen portions of objects due to its knowledge of 2D priors, it seems that ours often better (\emph{e.g.}, ABO –– note how EscherNet creates counterintuitive holes and Zero-123XL struggles to change viewpoints correctly).

\paragraph{Ablation}
We compare our model with a version that uses only global encodings, and one that keeps AdaLN with its classifier head. The results in Tab.~\ref{tab:ablation} show that both of the tested design choices were necessary for the model to achieve acceptable results.

\begin{figure}
\centering
\includegraphics[width=.45\textwidth]{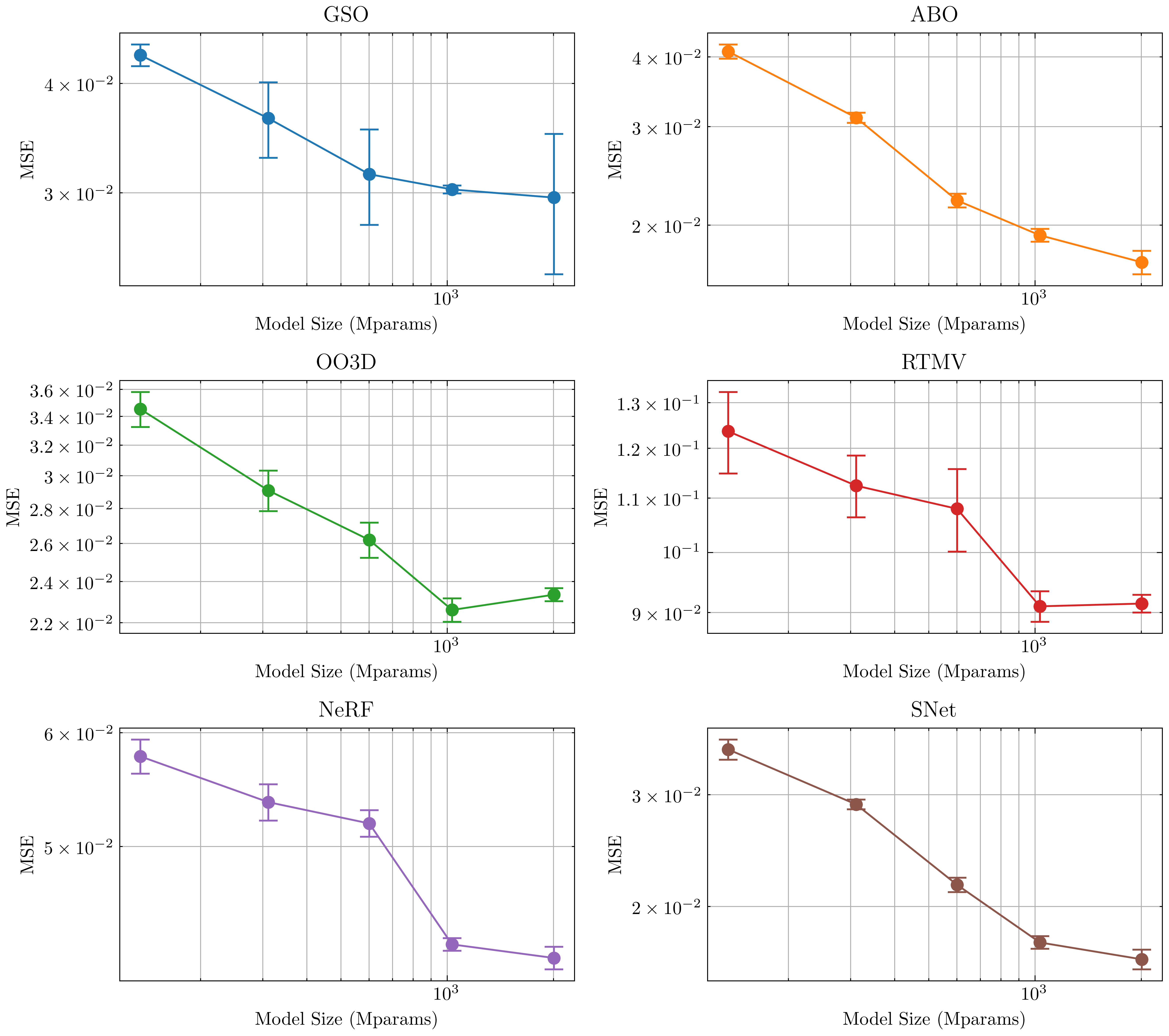}
\caption{\textbf{ArchonView scaling with model size.} The five data points respectively have depth 12, 16, 20, 24, and 30. Error bars are $\pm$ one standard deviation from five repetitions each.}
\label{fig:scaling_params}
\vspace{-2ex}
\end{figure}

\paragraph{Scaling}
We investigate the model's scaling behavior with respect to model size and dataset size (we do not consider scaling with computation due to the difficulty of rigorously defining an optimal stopping point, as in our case performance is not directly tied to token accuracy unlike in other tasks like LLMs). Results are shown in Figs.~\ref{fig:scaling_params} and~\ref{fig:scaling_data}. It seems that adding more data would likely continue to improve model performance considerably. As for model size, results demonstrate preliminary evidence of scaling law-like behavior below 1B parameters, and the performance of larger models is likely bottlenecked by dataset size.

\section{Discussion}

\subsection{Implications}
\begin{figure}
\centering
\includegraphics[width=.45\textwidth]{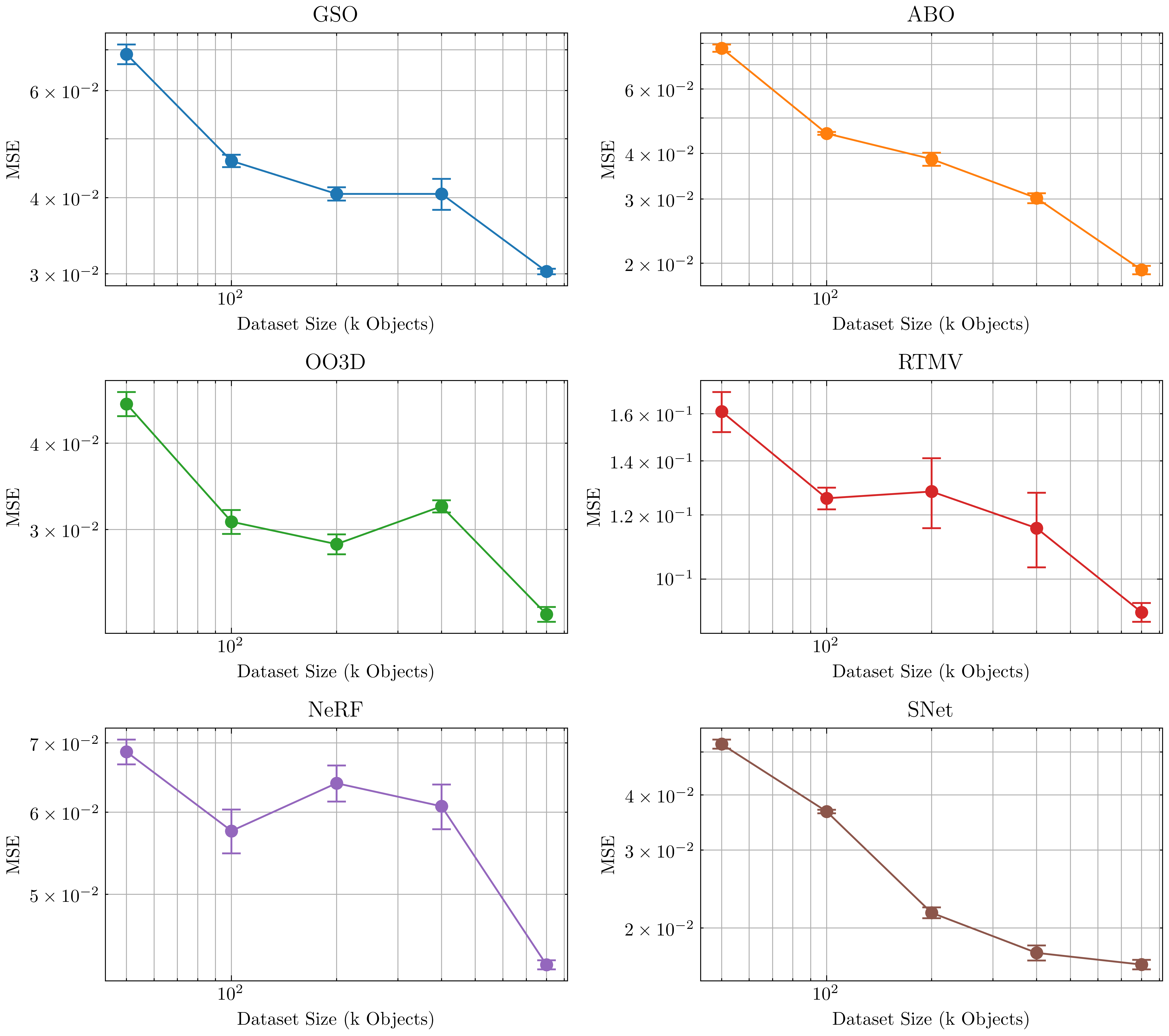}
\caption{\textbf{ArchonView scaling with dataset size.} We randomly subsample objects from our training set and train models with those subsets. Error bars are $\pm$ one standard deviation from five repetitions each.}
\label{fig:scaling_data}
\vspace{-2ex}
\end{figure}

According to~\cite{liu2023zero}, diffusion-based object-centric NVS models work by adapting visual priors inherent in pretrained 2D diffusion models. This, in turn, causes heavy reliance on large pretrained models such as Stable Diffusion (which was trained on over 2B images~\cite{schuhmann2022laion}). In contrast, we only use the ``fine-tuning'' dataset of previous works (with 800k 3D objects) and achieved significantly superior results. This demonstrates that under the autoregressive formulation, the 3D objects already contain enough information for the model to accurately, efficiently, and scalably conduct object-centric NVS. Furthermore, trends in Fig.~\ref{fig:scaling_data} show that increasing the dataset size beyond 800k (Objaverse) is likely to yield further positive results, which gives a clear direction for the next step of scaling ArchonView towards operational usage.

In addition, our superior speed and accuracy suggests the potential for a paradigm shift from diffusion to next-scale autoregression in the field of generative NVS, which avoids diffusion's ``original sins'' in speed and scalability that stem from the need for repeated denoising steps. Next-scale autoregression is also a rapidly evolving technique, and many advances in this line of research can possibly be used as plug-in improvements to our method. Hence we believe ArchonView can function as a base model for future innovations that can further drive this paradigm shift.

\subsection{Potential Negative Social Impacts}
Results presented by this work, given their visual generative nature, are prone to being exploited by malicious agents. We encourage responsible usage in accordance with relevant common guidelines. The ``Direct Use'' portion of the DALL-E Mini model card~\citep{dalle}, shared by works such as Stable Diffusion~\citep{rombach2022high}, applies well.

\section{Conclusion}
We introduce the first method of zero-shot single-image object-centric NVS to be based on visual autoregression, using the next-scale autoregression paradigm. We show that it does not require 2D pretraining, achieves state-of-the-art performance across several benchmark tasks, is several times faster in inference time compared to previous methods, and demonstrates scaling behavior with model size and dataset size. This demonstrates the oft-overlooked potential of autoregressive backbones and their advantages over diffusion for NVS tasks, and provides a base model for potential future work in this direction.

\section*{Acknowledgments}
We would like to thank Xiaoxue Chen (Tsinghua University) and Xin Kong (Imperial College London) for useful discussions. 

{
    \bibliographystyle{ieeenat_fullname}
    \bibliography{main}
}
\end{document}